\documentclass[sigconf,nonacm]{acmart}

\usepackage[utf8]{inputenc}
\usepackage[T1]{fontenc}
\usepackage{microtype}
\usepackage[ngerman,english]{babel}
\usepackage{listings} %
\usepackage{cleveref}
\usepackage{multicol}
\usepackage{glossaries}
\usepackage{tikz}
\usepackage{siunitx}
\usepackage{dblfloatfix}

\definecolor{color0}{HTML}{1f77b4}  %
\definecolor{color1}{HTML}{ff7f0e}  %
\definecolor{color2}{HTML}{2ca02c}  %
\definecolor{color3}{HTML}{d62728}  %
\definecolor{color4}{HTML}{9467bd}  %
\definecolor{color5}{HTML}{8c564b}  %
\definecolor{color6}{HTML}{e377c2}  %
\definecolor{color7}{HTML}{7f7f7f}  %
\definecolor{color8}{HTML}{bcbd22}  %
\definecolor{color9}{HTML}{17becf}  %

\usetikzlibrary{calc,backgrounds,fit,positioning}

\DeclareMathOperator*{\argmax}{\arg\!\max}

\newacronym{adc}{ADC}{analog-to-digital converter}
\newacronym{adex}{AdEx}{adaptive exponential integrate-and-fire}
\newacronym{afib}{AF}{atrial fibrillation}
\newacronym{ann}{ANN}{artificial neural network}
\newacronym{asic}{ASIC}{application-specific integrated circuit}
\newacronym{asicab}{\acrshort{asic} adapter \acrshort{pcb}}{\acrlong{asic} adapter \acrlong{pcb}}
\newacronym{api}{API}{application programming interface}
\newacronym{bmbf}{BMBF}{German Federal Ministry of Education and Research}
\newacronym{bptt}{BPTT}{backpropagation through time}
\newacronym{bss2}{\mbox{BSS-2}}{Brain\mbox{ScaleS-2}}
\newacronym{bss1}{\mbox{BSS-1}}{Brain\mbox{ScaleS-1}}
\newacronym{bss2os}{\gls{bss2} OS}{\gls{bss2} Operating System}
\newacronym{bss}{BSS}{BrainScaleS}
\newacronym{cdnn}{CDNN}{convolutional deep neural network}
\newacronym{cpu}{CPU}{central processing unit}
\newacronym{dfki}{DFKI}{German Research Centre for Artificial Intelligence}
\newacronym{dma}{DMA}{direct memory access}
\newacronym{dram}{DRAM}{dynamic random-access memory}
\newacronym{ecg}{ECG}{electrocardiogram}
\newacronym{fpga}{FPGA}{field-programmable gate array}
\newacronym{gbe}{GbE}{gigabit ethernet}
\newacronym{i2c}{I\textsuperscript{2}C}{Inter-Integrated Circuit}
\newacronym{ic}{IC}{integrated circuit}
\newacronym{isa}{ISA}{instruction set architecture}
\newacronym{itl}{ITL}{in-the-loop}
\newacronym{jit}{JIT}{just-in-time}
\newacronym{lvds}{LVDS}{low-voltage differential signaling}
\newacronym{lif}{LIF}{leaky-integrate and fire}
\newacronym{li}{LI}{leaky integrator}
\newacronym{mac}{MAC}{multiply–accumulate}
\newacronym{madc}{MADC}{membrane \acrshort{adc}}
\newacronym{mse}{MSE}{mean squared error}
\newacronym{cadc}{CADC}{columnar \acrshort{adc}}
\newacronym{pcb}{PCB}{printed circuit board}
\newacronym{ppu}{\acrshort{simd} \acrshort{cpu}}{\acrlong{simd} \acrlong{cpu}}
\newacronym{relu}{ReLU}{rectified linear unit}
\newacronym{rtl}{RTL}{Register Transfer Level}
\newacronym{gd}{GD}{gradient descent}
\newacronym{simd}{SIMD}{single instruction, multiple data}
\newacronym{snn}{SNN}{spiking neural network}
\newacronym{sodimm}{\mbox{SO-DIMM}}{small outline dual in-line memory module}
\newacronym{sram}{SRAM}{static random-access memory}
\newacronym{stdp}{STDP}{spike timing dependent plasticity}
\newacronym{stp}{STP}{short term plasticity}
\newacronym{rnn}{RNN}{recurrent neural network}
\newacronym{rsnn}{RSNN}{recurrent spiking neural network}
\newacronym{nasprop}{NASProp}{neuromorphic accumulative spike propagation}
\newacronym{vu}{VU}{vector unit}
\newacronym{udp}{UDP}{user datagram protocol}
\newacronym{cd}{CD}{continuous deployment}
\newacronym{ci}{CI}{continuous integration}
\newacronym{hpc}{HPC}{high-performance computing}
\newacronym{gpu}{GPU}{graphics processing unit}
\newacronym{usb}{USB}{universal serial bus}

\begin{document}

\title[Machine-learning-inspired Spiking Neural Network Modeling on BrainScaleS-2]{\texttt{hxtorch.snn}: Machine-learning-inspired Spiking Neural Network Modeling on BrainScaleS-2}

\author{Philipp Spilger}%
\authornote{Contributed equally.}%
\affiliation{%
	\institution{Kirchhoff-Institute for Physics, Heidelberg University}%
	\city{Heidelberg}%
	\country{Germany}%
}%
\email{pspilger@kip.uni-heidelberg.de}

\author{Elias Arnold}
\authornotemark[1]%
\affiliation{%
	\institution{Kirchhoff-Institute for Physics, Heidelberg University}
	\city{Heidelberg}
	\country{Germany}
}
\author{Luca Blessing}
\affiliation{%
	\institution{Kirchhoff-Institute for Physics, Heidelberg University}
	\city{Heidelberg}
	\country{Germany}
}
\author{Christian Mauch}
\affiliation{%
	\institution{Kirchhoff-Institute for Physics, Heidelberg University}
	\city{Heidelberg}
	\country{Germany}
}
\author{Christian Pehle}
\affiliation{%
	\institution{Kirchhoff-Institute for Physics, Heidelberg University}
	\city{Heidelberg}
	\country{Germany}
}
\author{Eric Müller}
\affiliation{%
	\institution{European Institute for Neuromorphic Computing, Heidelberg University}
	\city{Heidelberg}
	\country{Germany}
}
\author{Johannes Schemmel}
\affiliation{%
	\institution{Kirchhoff-Institute for Physics, Heidelberg University}
	\city{Heidelberg}
	\country{Germany}
}

\setcopyright{none}
\acmConference{}

\begin{abstract}
Neuromorphic systems require user-friendly software to support the design and optimization of experiments.
In this work, we address this need by presenting our development of a machine learning-based modeling framework for the \acrlong{bss2} neuromorphic system.
This work represents an improvement over previous efforts, which either focused on the matrix-multiplication mode of BrainScaleS-2 or lacked full automation.
Our framework, called \texttt{hxtorch.snn}, enables the hardware-\acrlong{itl} training of \acrlongpl{snn} within PyTorch, including support for auto differentiation in a fully-automated hardware experiment workflow.
In addition, \texttt{hxtorch.snn} facilitates seamless transitions between emulating on hardware and simulating in software.
We demonstrate the capabilities of \texttt{hxtorch.snn} on a classification task using the Yin-Yang dataset employing a gradient-based approach with surrogate gradients and densely sampled membrane observations from the \acrlong{bss2} hardware system.

\end{abstract}

\maketitle

\section{Introduction}\label{sec:introduction}

Modern \gls{hpc} environments speed up individual workloads with domain-specific hardware accelerators~\cite{dally2020domainspecific}.
However, successfully establishing novel computing par\-a\-digms depends as much on the availability of a computing substrate as it does on the software infrastructure that provides the means to work with that substrate.
In particular, the `programming model' of neuromorphic hardware is a very different approach to data processing than conventional systems~\cite{mueller2022operating,mueller2022scalable}.
Recently, machine-learning-inspired training methods, and in particular gradient-based optimization of \glspl{snn}, have become increasingly popular~\cite{neftci2019surrogate, bellec2018long, shrestha2018slayer, esser2016convolutional, esser2015backpropagation}.

We introduce \texttt{hxtorch.snn}, a PyTorch~\cite{paszke2019pytorch} wrapper library for the accelerated mixed-signal neuromorphic hardware system \gls{bss2}~\cite{pehle2022brainscales2}.
We discuss and demonstrate our implementation using a simple benchmark task trained with a gradient-based method and the \gls{bss2} hardware system in the loop.
The implementation described here has also been used in a real-world application \cite{arnold2022spikinghardware} and will serve as the basis for future experiments.

Here we use the gradient estimation method introduced in \cite{cramer2022surrogate}, which uses the unrolled computational graph determined by the forward pass, to perform gradient estimation by injecting hardware voltage measurements and spikes in the backward pass.
The overall library, however, is agnostic to this particular choice of gradient estimation method.
Other approaches can be implemented, such as using a closed-form analytical formula \cite{goeltz2021fast} or the EventProp algorithm \cite{wunderlich2021event}.
To support this generic framework for gradient estimation, we implement utilities for interpolation and normalization of hardware measurements, conversion of spike observations into dense tensors, and procedures for weight quantization. 

Existing approaches to interfacing with Neuromorphic hardware tend to favor abstractions familiar to neuro-scientists, such as populations and projections \cite{rhodes2018spynnaker,rowley2019spinntools, lin2018programming,davison2009pynn,bekolay2014nengo} and only a few supported gradient-based methods \cite{hananel2018BindsNet,rockpool2019docs}.
Recently several libraries for machine learning with spiking neurons in PyTorch have appeared, among them, Norse~\cite{pehle2021norse}, Lava~\cite{loihi2021lava_misc}, NxTF~\cite{rueckauer2021nxtf}, snnTorch~\cite{eshraghian2021training}, and SpikingJelly~\cite{SpikingJelly}.

The \gls{bss2} mixed-signal neuromorphic hardware emulates networks of spiking neurons time-continuously in analog circuits~\cite{pehle2022brainscales2}.
Each of the 512 \gls{adex}~\cite{brette2005adaptive} neuron compartments on \gls{bss2} are individually parameterized and, in particular, can be configured to resemble \gls{lif} and \gls{li} neuron dynamics.
They receive input stimuli from a column of 256 exponential synapses in the synapse matrix with adjustable 6-bit weights.
Configuration of the on-chip routing mechanism and the synapse matrix allows addressing the neurons' digital spike events to target synapses and, thus, the realization of arbitrary topologies.
In addition, \glspl{ann} can be emulated by using the neurons in a non-spiking mode, which allows implementing analog vector-matrix multiplications~\cite{weis2020inference,stradmann2022demonstrating}.
The recordings of spike events and membrane potentials, sampled in parallel via a \gls{cadc}, are accessible by the host computer and can be incorporated, e.g., for weight update computation in a hardware-\gls{itl}~\cite{schmitt2017hwitl} fashion.
A \gls{fpga} serves as the real-time experiment master and as interconnect between the neuromorphic chip and a host computer.

We use a layered software approach to abstract hardware usage~\cite{mueller2022scalable}.
At the lowest level, a transport layer and an instruction set of commands to the \gls{fpga}, which execute sequentially, are employed.
Above, individual hardware entity configuration, e.g., for each neuron, is abstracted into containers residing in a uniform address space for type safety and to remove requiring knowledge about the memory layout.
Using this hardware abstraction, a data-flow-graph-based experiment notation, \texttt{grenade}, separates network topology specification, temporal evolution description, and execution to support the accelerated real-time nature of the neuromorphic hardware.
On top, front ends for spiking and non-spiking experiments exist.
Spiking neural networks can be described in the back-end-agnostic language \texttt{PyNN}~\cite{davison2009pynn} targeting computational neuroscience.
For offloading \glspl{ann}, i.e.\ analog vector-matrix multiplications, a thin PyTorch machine-learning adapter, \texttt{hxtorch}, was developed~\cite{spilger2020hxtorch}, separating hardware interaction and machine-learning framework.
Similarly, this work adds support for describing \glspl{snn} in this machine-learning framework yielding access to training, inference, and data alteration integrated into PyTorch's ecosystem.

\section{Methods}\label{sec:methods}

For developing a machine-learning framework supporting the emulation of \glspl{snn} on \gls{bss2}, we follow the same structure as for the developed \gls{ann} machine-learning adaptor \texttt{hxtorch}~\cite{spilger2020hxtorch}.
Due to its vast community, we continue to build our front end upon the PyTorch framework.
Given the accelerated time-continuous nature of the neuromorphic hardware, network topology specification, temporal evolution and execution need to be separated in the machine-learning front end.
In particular, individual numerical operations in a time-grid-based simulation cannot be trivially identified with dynamics on the neuromorphic substrate.

We use the data-flow-graph-based experiment notation~\cite{mueller2022scalable} as the back end.
The machine-learning front end then needs to handle data conversion to and from PyTorch tensors and wrapping of network topology descriptions in a PyTorch-compatible notation.
Therefore, hardware interaction and machine-learning framework adaptation are separated.

In PyTorch, models are eagerly executed by the computation of a layer's result when calling its \texttt{forward} method, which implies that the network builds up incrementally.
However, the emulation of \glspl{snn} in physical time requires knowledge of the complete network topology before experiment execution, to derive a corresponding hardware configuration.
Therefore, an additional separation is required.

Since training neural networks benefits from the auto differentiation provided by the machine-learning framework, we aim to incorporate this into the network description.
Within PyTorch, we train with the hardware in the loop by emulating the forward pass on \gls{bss2} and injecting its results in the backward pass on the host computer.
Moreover, the convenience of describing models by a composition of modules, i.e.\ network layers, should be maintained and allows for defining network topologies with recurrently connected layers.

For developing a model for the neuromorphic hardware from scratch or a reference model, iterative development is beneficial.
Exchanging a simulated model with emulation on \gls{bss2} requires parameter translations, e.g., weight scaling and adaptation to limited resolution on hardware or neuron model parameter translation to technical hardware parameters.
A seamless comparison to and replacement of (parts of) the network by (custom) simulation facilitates this.
It also enables hybrid networks, for example, by adding the possibility to introduce layers not representable on the neuromorphic hardware, e.g., neuron models not present on hardware or other arbitrary numerical operations on intermediate results, while retaining the neuromorphic hardware's acceleration for the representable parts.

Lastly, the collection of hardware observables, such as membrane potential recordings or spike trains, shall be adjustable per layer depending on the training algorithms.
Importantly, the transformation of acquired hardware observables into PyTorch data structures needs to be fully customizable to be best-tailored for the training algorithm at hand, e.g., the transition of sparse time-series data to a dense time grid, the interpolation of sparse membrane measurements,
or the mapping of synapse weight parameters between PyTorch and \gls{bss2} can impact the network's dynamics as well as the learning process.

\section{Results}\label{sec:results}
The software library \texttt{hxtorch.snn} integrates modeling \glspl{snn} on \gls{bss2} into the PyTorch ecosystem.
It extends the thin wrapper library \texttt{hxtorch}, which previously only targeted modeling \glspl{ann}.

\Cref{fig:swflow} shows the library structure, experiment description, and constitution.
\begin{figure*}[t]
	\centering
	\begin{minipage}[b]{0.59\linewidth}
		\textbf{A}
	\end{minipage}
	\begin{minipage}[b]{0.39\linewidth}
		\textbf{B}
	\end{minipage}
	\begin{minipage}[b]{0.59\linewidth}
		\input{swflow.tex}
	\end{minipage}
	\begin{minipage}[b]{0.39\linewidth}
		\begin{lstlisting}[basicstyle=\tt\small, escapeinside=\'\']
'\tt\color{color0}ins' = hxtorch.snn.'\tt\color{color0}Instance'()

'\tt\color{color1}syn1' = hxtorch.snn.'\tt\color{color1}Synapse'('\tt\color{color0}ins', ...)
'\tt\color{color1}lif1' = hxtorch.snn.'\tt\color{color1}LIF'('\tt\color{color0}ins', ...)
'\tt\color{color1}syn2' = hxtorch.snn.'\tt\color{color1}Synapse'('\tt\color{color0}ins', ...)
'\tt\color{color1}li2' = hxtorch.snn.'\tt\color{color1}LI'('\tt\color{color0}ins', ...)

'\tt\color{color2}x1' = '\tt\color{color1}syn1'('\tt\color{color2}input')
'\tt\color{color2}x2' = '\tt\color{color1}lif1'('\tt\color{color2}x1')
'\tt\color{color2}x3' = '\tt\color{color1}syn2'('\tt\color{color2}x2')
'\tt\color{color2}x4' = '\tt\color{color1}li2'('\tt\color{color2}x3')

hxtorch.snn.'\tt\color{color0}run'('\tt\color{color0}ins', ...)

'\tt\color{color6}loss' = f('\tt\color{color2}x4')
'\tt\color{color6}loss'.backward()
		\end{lstlisting}
	\end{minipage}
	\caption{\label{fig:swflow}
		Software \gls{api} of \texttt{hxtorch.snn} shown exemplarily for the feed-forward network used in the classification of the Yin-Yang dataset~\cite{kriener2021yin}, where \textbf{A} depicts the data flow in the model and with \gls{bss2} and \textbf{B} displays the corresponding source code.
		The network consists of two synapse layers and two \gls{li}/\gls{lif} neuron layers and is associated to an instance to be executed on hardware.
		Execution is triggered explicitly via the \texttt{run} function and the handles \texttt{x}, \texttt{y} and \texttt{loss} only afterwards carry their result data from the execution.
		This then in particular allows gradient calculation via backward functions attached to the handles' data.
	}
\end{figure*}
Similarly to \gls{ann} models in PyTorch, we describe constituents of a \gls{snn} model like synapse and neuron layers via custom modules derived from \texttt{HXModule}, containing parameters and representing the corresponding hardware entity.
We want to retain PyTorch's eager model construction and execution interface by successively applying modules' \texttt{forward(input)} calls, but we need to separate the construction of the complete model from its execution on the hardware.
Therefore, we only register each module's invocation into an \texttt{Instance}, passed to each module on its construction.
Instead of returning the actual results, we return \texttt{Handle}-typed promises of future results.
The instance is then separately executable, and the promises are filled with the result data of the respective modules after the instance's execution.
An instance represents one hardware experiment execution.
Upon explicit execution of an instance via \texttt{run}, the network topology is extracted from the registered modules and their invocations and converted into a graph-based description, which is then mapped to a hardware experiment description and executed on hardware.
The resulting data is transformed into PyTorch tensors, post-processed via a customizable method, and finally annotated onto the corresponding data handles, thereby being accessible by the user.
Since instances of \texttt{Instance} only require hardware allocations upon execution, multiple instances can coexist and be executed sequentially.
This allows interleaving hardware-executed model parts and simulated parts as long as no inter-instance recurrence is required.%

The model construction via network entity module invocations allows defining recurrent models by associating each module instance with a network entity and using invocations to connect them, cf.\ \cref{lst:recurrence}.
\begin{lstlisting}[basicstyle=\tt\small, caption={\label{lst:recurrence}Construction of a recurrent network. When reusing the same network entity \texttt{nrn}, a recurrent connection is created.}]
nrn = hxtorch.snn.LIF(...)
syn1 = hxtorch.snn.Synapse(...)
syn2 = hxtorch.snn.Synapse(...)

x = syn1(input)
x = nrn(x)  # feed-forward
x = syn2(x)
x = nrn(x)  # recurrence
\end{lstlisting}

Upon construction, each module is equipped with a custom PyTorch-differentiable
function, either directly as \texttt{Auto\-grad.Func\-tion} or via a function defining both a simulated forward pass and an implicit backward pass.
In the first case, \texttt{hxtorch.snn} injects the hardware data implicitly to be used in the backward function, and in the second case, hardware observables are provided as an additional function argument.
This allows annotating a backward pass to the handle data tensors such that PyTorch's auto differentiation can seamlessly backpropagate gradients based on hardware observations.
When a simulated forward pass is provided, this module can also be used without hardware usage as long as it does not participate in a cyclic sub-network.
This greatly aids network development and translation onto the hardware, since (parts of) a simulated reference network can gradually be interchanged with and compared to hardware entities within the same library.

Currently, \texttt{hxtorch.snn} supports \gls{lif} and \gls{li} neuron layers and has access to spike times and membrane voltage recordings of individual hardware neurons.
To support the full extent of \gls{bss2} neuron dynamics, a forward and backward PyTorch implementation of the \gls{adex} neuron model is planned.

A dropout module provides additional machine learning functionality by applying a batch-wise spiking mask to a preceding neuron layer and disabling the spike output on hardware accordingly.

Synaptic connections on \gls{bss2} are exposed to PyTorch as a \texttt{Synapse} module, representing a projection between neuron layers.
This provides a framework for recording of additional synapse-related observables, such as spike-time correlation sensor recordings.

\begin{figure*}[tb]
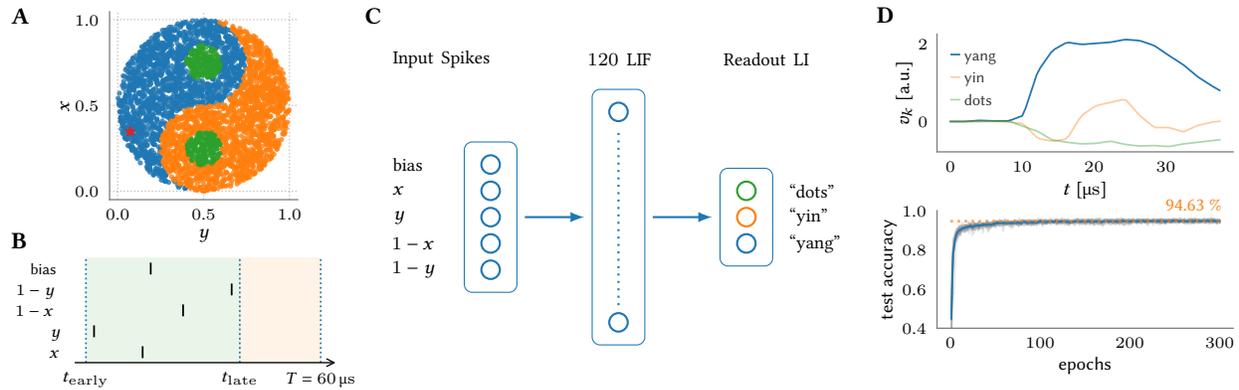

    \centering
    \tikzset{
        panel/.style={
            inner sep=0pt,
            outer sep=0pt,
			execute at begin node={\tikzset{anchor=center, inner sep=.33333emh}}},
        label/.style={
            anchor=north west,
            inner sep=0,
            outer sep=0}}

    \begin{tikzpicture}
        \node[panel, anchor=north west] (b) at (0, -3) {
			%% Creator: Matplotlib, PGF backend
%%
%% To include the figure in your LaTeX document, write
%%   \input{<filename>.pgf}
%%
%% Make sure the required packages are loaded in your preamble
%%   \usepackage{pgf}
%%
%% Also ensure that all the required font packages are loaded; for instance,
%% the lmodern package is sometimes necessary when using math font.
%%   \usepackage{lmodern}
%%
%% Figures using additional raster images can only be included by \input if
%% they are in the same directory as the main LaTeX file. For loading figures
%% from other directories you can use the `import` package
%%   \usepackage{import}
%%
%% and then include the figures with
%%   \import{<path to file>}{<filename>.pgf}
%%
%% Matplotlib used the following preamble
%%   \usepackage[utf8x]{inputenc}\usepackage[T1]{fontenc}\usepackage[detect-all,binary-units]{siunitx}\DeclareSIUnit\operation{op}
%%
\begingroup%
\makeatletter%
\begin{pgfpicture}%
\pgfpathrectangle{\pgfpointorigin}{\pgfqpoint{1.839749in}{0.830000in}}%
\pgfusepath{use as bounding box, clip}%
\begin{pgfscope}%
\pgfsetbuttcap%
\pgfsetmiterjoin%
\definecolor{currentfill}{rgb}{1.000000,1.000000,1.000000}%
\pgfsetfillcolor{currentfill}%
\pgfsetlinewidth{0.000000pt}%
\definecolor{currentstroke}{rgb}{1.000000,1.000000,1.000000}%
\pgfsetstrokecolor{currentstroke}%
\pgfsetdash{}{0pt}%
\pgfpathmoveto{\pgfqpoint{0.000000in}{0.000000in}}%
\pgfpathlineto{\pgfqpoint{1.839749in}{0.000000in}}%
\pgfpathlineto{\pgfqpoint{1.839749in}{0.830000in}}%
\pgfpathlineto{\pgfqpoint{0.000000in}{0.830000in}}%
\pgfpathlineto{\pgfqpoint{0.000000in}{0.000000in}}%
\pgfpathclose%
\pgfusepath{fill}%
\end{pgfscope}%
\begin{pgfscope}%
\pgfsetbuttcap%
\pgfsetmiterjoin%
\definecolor{currentfill}{rgb}{1.000000,1.000000,1.000000}%
\pgfsetfillcolor{currentfill}%
\pgfsetlinewidth{0.000000pt}%
\definecolor{currentstroke}{rgb}{0.000000,0.000000,0.000000}%
\pgfsetstrokecolor{currentstroke}%
\pgfsetstrokeopacity{0.000000}%
\pgfsetdash{}{0pt}%
\pgfpathmoveto{\pgfqpoint{0.305266in}{0.152892in}}%
\pgfpathlineto{\pgfqpoint{1.724628in}{0.152892in}}%
\pgfpathlineto{\pgfqpoint{1.724628in}{0.810000in}}%
\pgfpathlineto{\pgfqpoint{0.305266in}{0.810000in}}%
\pgfpathlineto{\pgfqpoint{0.305266in}{0.152892in}}%
\pgfpathclose%
\pgfusepath{fill}%
\end{pgfscope}%
\begin{pgfscope}%
\definecolor{textcolor}{rgb}{0.000000,0.000000,0.000000}%
\pgfsetstrokecolor{textcolor}%
\pgfsetfillcolor{textcolor}%
\pgftext[x=0.390004in,y=0.104281in,,top]{\color{textcolor}\sffamily\fontsize{6.000000}{7.200000}\selectfont \(\displaystyle t_\mathrm{early}\)}%
\end{pgfscope}%
\begin{pgfscope}%
\definecolor{textcolor}{rgb}{0.000000,0.000000,0.000000}%
\pgfsetstrokecolor{textcolor}%
\pgfsetfillcolor{textcolor}%
\pgftext[x=1.195016in,y=0.104281in,,top]{\color{textcolor}\sffamily\fontsize{6.000000}{7.200000}\selectfont \(\displaystyle t_\mathrm{late}\)}%
\end{pgfscope}%
\begin{pgfscope}%
\definecolor{textcolor}{rgb}{0.000000,0.000000,0.000000}%
\pgfsetstrokecolor{textcolor}%
\pgfsetfillcolor{textcolor}%
\pgftext[x=1.618706in,y=0.104281in,,top]{\color{textcolor}\sffamily\fontsize{6.000000}{7.200000}\selectfont \(\displaystyle T=\SI{60}{\micro\second}\)}%
\end{pgfscope}%
\begin{pgfscope}%
\definecolor{textcolor}{rgb}{0.000000,0.000000,0.000000}%
\pgfsetstrokecolor{textcolor}%
\pgfsetfillcolor{textcolor}%
\pgftext[x=0.198496in, y=0.177466in, left, base]{\color{textcolor}\sffamily\fontsize{6.000000}{7.200000}\selectfont \(\displaystyle x\)}%
\end{pgfscope}%
\begin{pgfscope}%
\definecolor{textcolor}{rgb}{0.000000,0.000000,0.000000}%
\pgfsetstrokecolor{textcolor}%
\pgfsetfillcolor{textcolor}%
\pgftext[x=0.200907in, y=0.286986in, left, base]{\color{textcolor}\sffamily\fontsize{6.000000}{7.200000}\selectfont \(\displaystyle y\)}%
\end{pgfscope}%
\begin{pgfscope}%
\definecolor{textcolor}{rgb}{0.000000,0.000000,0.000000}%
\pgfsetstrokecolor{textcolor}%
\pgfsetfillcolor{textcolor}%
\pgftext[x=0.020000in, y=0.396502in, left, base]{\color{textcolor}\sffamily\fontsize{6.000000}{7.200000}\selectfont \(\displaystyle 1-x\)}%
\end{pgfscope}%
\begin{pgfscope}%
\definecolor{textcolor}{rgb}{0.000000,0.000000,0.000000}%
\pgfsetstrokecolor{textcolor}%
\pgfsetfillcolor{textcolor}%
\pgftext[x=0.022411in, y=0.506022in, left, base]{\color{textcolor}\sffamily\fontsize{6.000000}{7.200000}\selectfont \(\displaystyle 1-y\)}%
\end{pgfscope}%
\begin{pgfscope}%
\definecolor{textcolor}{rgb}{0.000000,0.000000,0.000000}%
\pgfsetstrokecolor{textcolor}%
\pgfsetfillcolor{textcolor}%
\pgftext[x=0.098739in, y=0.615538in, left, base]{\color{textcolor}\sffamily\fontsize{6.000000}{7.200000}\selectfont bias}%
\end{pgfscope}%
\begin{pgfscope}%
\pgfpathrectangle{\pgfqpoint{0.305266in}{0.152892in}}{\pgfqpoint{1.419362in}{0.657108in}}%
\pgfusepath{clip}%
\pgfsetbuttcap%
\pgfsetmiterjoin%
\definecolor{currentfill}{rgb}{1.000000,0.498039,0.054902}%
\pgfsetfillcolor{currentfill}%
\pgfsetfillopacity{0.100000}%
\pgfsetlinewidth{0.501875pt}%
\definecolor{currentstroke}{rgb}{1.000000,0.498039,0.054902}%
\pgfsetstrokecolor{currentstroke}%
\pgfsetstrokeopacity{0.100000}%
\pgfsetdash{}{0pt}%
\pgfpathmoveto{\pgfqpoint{1.195016in}{0.152892in}}%
\pgfpathlineto{\pgfqpoint{1.618706in}{0.152892in}}%
\pgfpathlineto{\pgfqpoint{1.618706in}{0.700482in}}%
\pgfpathlineto{\pgfqpoint{1.195016in}{0.700482in}}%
\pgfpathlineto{\pgfqpoint{1.195016in}{0.152892in}}%
\pgfpathclose%
\pgfusepath{stroke,fill}%
\end{pgfscope}%
\begin{pgfscope}%
\pgfpathrectangle{\pgfqpoint{0.305266in}{0.152892in}}{\pgfqpoint{1.419362in}{0.657108in}}%
\pgfusepath{clip}%
\pgfsetbuttcap%
\pgfsetmiterjoin%
\definecolor{currentfill}{rgb}{0.172549,0.627451,0.172549}%
\pgfsetfillcolor{currentfill}%
\pgfsetfillopacity{0.100000}%
\pgfsetlinewidth{0.501875pt}%
\definecolor{currentstroke}{rgb}{0.172549,0.627451,0.172549}%
\pgfsetstrokecolor{currentstroke}%
\pgfsetstrokeopacity{0.100000}%
\pgfsetdash{}{0pt}%
\pgfpathmoveto{\pgfqpoint{0.390004in}{0.152892in}}%
\pgfpathlineto{\pgfqpoint{1.195016in}{0.152892in}}%
\pgfpathlineto{\pgfqpoint{1.195016in}{0.700482in}}%
\pgfpathlineto{\pgfqpoint{0.390004in}{0.700482in}}%
\pgfpathlineto{\pgfqpoint{0.390004in}{0.152892in}}%
\pgfpathclose%
\pgfusepath{stroke,fill}%
\end{pgfscope}%
\begin{pgfscope}%
\pgfpathrectangle{\pgfqpoint{0.305266in}{0.152892in}}{\pgfqpoint{1.419362in}{0.657108in}}%
\pgfusepath{clip}%
\pgfsetbuttcap%
\pgfsetroundjoin%
\pgfsetlinewidth{0.662475pt}%
\definecolor{currentstroke}{rgb}{0.121569,0.466667,0.705882}%
\pgfsetstrokecolor{currentstroke}%
\pgfsetdash{{0.660000pt}{1.089000pt}}{0.000000pt}%
\pgfpathmoveto{\pgfqpoint{0.390004in}{0.152892in}}%
\pgfpathlineto{\pgfqpoint{0.390004in}{0.700482in}}%
\pgfusepath{stroke}%
\end{pgfscope}%
\begin{pgfscope}%
\pgfpathrectangle{\pgfqpoint{0.305266in}{0.152892in}}{\pgfqpoint{1.419362in}{0.657108in}}%
\pgfusepath{clip}%
\pgfsetbuttcap%
\pgfsetroundjoin%
\pgfsetlinewidth{0.662475pt}%
\definecolor{currentstroke}{rgb}{0.121569,0.466667,0.705882}%
\pgfsetstrokecolor{currentstroke}%
\pgfsetdash{{0.660000pt}{1.089000pt}}{0.000000pt}%
\pgfpathmoveto{\pgfqpoint{1.195016in}{0.152892in}}%
\pgfpathlineto{\pgfqpoint{1.195016in}{0.700482in}}%
\pgfusepath{stroke}%
\end{pgfscope}%
\begin{pgfscope}%
\pgfpathrectangle{\pgfqpoint{0.305266in}{0.152892in}}{\pgfqpoint{1.419362in}{0.657108in}}%
\pgfusepath{clip}%
\pgfsetbuttcap%
\pgfsetroundjoin%
\pgfsetlinewidth{0.662475pt}%
\definecolor{currentstroke}{rgb}{0.121569,0.466667,0.705882}%
\pgfsetstrokecolor{currentstroke}%
\pgfsetdash{{0.660000pt}{1.089000pt}}{0.000000pt}%
\pgfpathmoveto{\pgfqpoint{1.618706in}{0.152892in}}%
\pgfpathlineto{\pgfqpoint{1.618706in}{0.700482in}}%
\pgfusepath{stroke}%
\end{pgfscope}%
\begin{pgfscope}%
\pgfsetroundcap%
\pgfsetroundjoin%
\pgfsetlinewidth{0.501875pt}%
\definecolor{currentstroke}{rgb}{0.000000,0.000000,0.000000}%
\pgfsetstrokecolor{currentstroke}%
\pgfsetdash{}{0pt}%
\pgfpathmoveto{\pgfqpoint{0.333031in}{0.152892in}}%
\pgfpathquadraticcurveto{\pgfqpoint{1.014922in}{0.152892in}}{\pgfqpoint{1.689048in}{0.152892in}}%
\pgfusepath{stroke}%
\end{pgfscope}%
\begin{pgfscope}%
\pgfsetroundcap%
\pgfsetroundjoin%
\pgfsetlinewidth{0.501875pt}%
\definecolor{currentstroke}{rgb}{0.000000,0.000000,0.000000}%
\pgfsetstrokecolor{currentstroke}%
\pgfsetdash{}{0pt}%
\pgfpathmoveto{\pgfqpoint{1.650159in}{0.172337in}}%
\pgfpathlineto{\pgfqpoint{1.689048in}{0.152892in}}%
\pgfpathlineto{\pgfqpoint{1.650159in}{0.133448in}}%
\pgfusepath{stroke}%
\end{pgfscope}%
\begin{pgfscope}%
\pgfpathrectangle{\pgfqpoint{0.305266in}{0.152892in}}{\pgfqpoint{1.419362in}{0.657108in}}%
\pgfusepath{clip}%
\pgfsetbuttcap%
\pgfsetroundjoin%
\definecolor{currentfill}{rgb}{0.000000,0.000000,0.000000}%
\pgfsetfillcolor{currentfill}%
\pgfsetlinewidth{0.662475pt}%
\definecolor{currentstroke}{rgb}{0.000000,0.000000,0.000000}%
\pgfsetstrokecolor{currentstroke}%
\pgfsetdash{}{0pt}%
\pgfsys@defobject{currentmarker}{\pgfqpoint{0.000000in}{-0.031056in}}{\pgfqpoint{0.000000in}{0.031056in}}{%
\pgfpathmoveto{\pgfqpoint{0.000000in}{-0.031056in}}%
\pgfpathlineto{\pgfqpoint{0.000000in}{0.031056in}}%
\pgfusepath{stroke,fill}%
}%
\begin{pgfscope}%
\pgfsys@transformshift{0.432373in}{0.317169in}%
\pgfsys@useobject{currentmarker}{}%
\end{pgfscope}%
\begin{pgfscope}%
\pgfsys@transformshift{0.686587in}{0.207651in}%
\pgfsys@useobject{currentmarker}{}%
\end{pgfscope}%
\begin{pgfscope}%
\pgfsys@transformshift{0.728956in}{0.645723in}%
\pgfsys@useobject{currentmarker}{}%
\end{pgfscope}%
\begin{pgfscope}%
\pgfsys@transformshift{0.898433in}{0.426687in}%
\pgfsys@useobject{currentmarker}{}%
\end{pgfscope}%
\begin{pgfscope}%
\pgfsys@transformshift{1.152647in}{0.536205in}%
\pgfsys@useobject{currentmarker}{}%
\end{pgfscope}%
\end{pgfscope}%
\end{pgfpicture}%
\makeatother%
\endgroup%
};
        \node[label] at (0, -3) {\textbf{B}};

        \node[panel, anchor=north west] (a) at (0.5,  0) {
			\input{yinyang_dataset.pgf}};
        \node[label] at (0, 0) {\textbf{A}};

        \node[panel, anchor=north west] (c) at (5.0,  -0.5) {
			\begin{tikzpicture}[
        neuron_i/.style={
            circle,
            inner sep=0pt,
            outer sep=3pt,
            align=center,
            thick,
            minimum size=7pt},
        neuron_t/.style={
            rectangle,
            inner sep=2pt,
            outer sep=3pt,
            align=left,
            thick,
	    text width=1cm,
            minimum width=1cm},
	outer/.style={
            rectangle,
            inner sep=4pt,
            outer sep=3pt,
            rounded corners=3pt,
            align=center}
]

\def\spacing{0.35}

% Input neurons
\node[neuron_i, draw=color0] (n_i_0) at (0, -2.*\spacing) {};
\node[neuron_i, draw=color0] (n_i_1) at (0, -1.*\spacing) {};
\node[neuron_i, draw=color0] (n_i_2) at (0, 0.*\spacing) {};
\node[neuron_i, draw=color0] (n_i_3) at (0, 1.*\spacing) {};
\node[neuron_i, draw=color0] (n_i_4) at (0, 2.*\spacing) {};
\node[outer, draw=color0, minimum height=2cm, minimum width=0.7cm] (n_i_o) at (0, 0) {};

\node[neuron_t, draw=none] (n_i_0_t) at (-0.8, 2.*\spacing) {\footnotesize bias};
\node[neuron_t, draw=none] (n_i_1_t) at (-0.8, 1.*\spacing) {\footnotesize $x$};
\node[neuron_t, draw=none] (n_i_2_t) at (-0.8, 0.*\spacing) {\footnotesize$y$};
\node[neuron_t, draw=none] (n_i_3_t) at (-0.8, -1.*\spacing) {\footnotesize$1-x$};
\node[neuron_t, draw=none] (n_i_4_t) at (-0.8, -2.*\spacing) {\footnotesize$1-y$};
\node[neuron_t, draw=none, text width=3cm, minimum width=3cm] (n_i_t) at (0.2, 6.*\spacing) {\footnotesize Input Spikes};

% Hidden neurons
\node[neuron_i, draw=color0] (n_h_0) at (1.7, -4.*\spacing) {};
\node[neuron_i, draw=color0] (n_h_1) at (1.7, 4.*\spacing) {};
\node[outer, draw=color0, minimum height=3.4cm, minimum width=0.7cm] (n_h_o) at (1.7, 0) {};
\draw[thick, draw=color0, dotted] (n_h_0.north) -- (n_h_1.south);
\node[neuron_t, draw=none, text width=3cm, minimum width=3cm] (n_h_t) at (2.8, 6.*\spacing) {\footnotesize $120$ LIF};

% Output neurons
\node[neuron_i, draw=color0] (n_o_0) at (3.4, -1.*\spacing) {};
\node[neuron_i, draw=color1] (n_o_1) at (3.4, 0.*\spacing) {};
\node[neuron_i, draw=color2] (n_o_2) at (3.4, 1.*\spacing) {};
\node[outer, draw=color0, minimum height=1.3cm, minimum width=0.7cm] (n_o_o) at (3.4, 0) {};
\node[neuron_t, draw=none, text width=3cm, minimum width=3cm] (n_h_t) at (4.6, 6.*\spacing) {\footnotesize Readout LI};
\node[neuron_t, draw=none] (n_o_0_t) at (4.5, -1.*\spacing) {\footnotesize``yang''};
\node[neuron_t, draw=none] (n_o_1_t) at (4.5, 0.*\spacing) {\footnotesize``yin''};
\node[neuron_t, draw=none] (n_o_2_t) at (4.5, 1.*\spacing) {\footnotesize``dots''};

% weights
\draw[thick, draw=color0, -latex] (n_i_o.east) -- (n_h_o.west);
\draw[thick, draw=color0, -latex] (n_h_o.east) -- (n_o_o.west);

\end{tikzpicture}};
        \node[label] at (4.7, 0) {\textbf{C}};

        \node[panel, anchor=north west] (d) at (11.5, -0.3) {
			\input{yinyang_loss.pgf}};
        \node[label] at (11.5, 0) {\textbf{D}};
    \end{tikzpicture}
	\caption{
		(A) Example samples from the Yin-Yang dataset \cite{kriener2021yin}.
		The dataset consists of samples defined on the $xy$-plane, each assigned to one of three nonlinear separable classes yin (orange), yang (blue), and dots (green).
		(B) The spike encoding of the 2D point from the Yin-Yang dataset in A depicted as a red star.
		Each dimension of the point and an inverse of it is translated to a spike event.
		Additionally, a bias spike is added to increase network activity.
		(C) The \acrshort{snn} topology used to classify the Yin-Yang dataset.
		An input layer projects the spike-encoded point onto a hidden layer consisting of \acrshort{lif} neurons.
		A \acrshort{li} readout layer receives the spike events of the hidden layer.
		Their maximal membrane values over time are used to infer a decision.
		(D) The upper plot depicts the analog output membrane voltages while inferring the red sample in A.
		In the lower plot the classification accuracy is shown over the test epochs.
		On \acrshort{bss2}, the \acrshort{snn} achieves an accuracy of $94.63 \pm 0.7$\%.
		The code is available as an interactive demo~\cite{brainscales2022yinyang}.
	}
	\label{fig:yinyang}
\end{figure*}

As a demonstration of our framework we consider the low-dimensional Yin-Yang classification task~\cite{kriener2021yin} visualized in \Cref{fig:yinyang}A.
It consists of three classes \emph{yin}, \emph{yang}, and \emph{dots}, each defining an area on the 2-dimensional $xy$-plane.
Its $i$-th sample is a point $\boldsymbol{x}_i = (x_i,y_i)$, randomly drawn from the corresponding areas and labeled accordingly.
As input to the \gls{snn}, the sample $\boldsymbol{x}_i$ is translated to spike times of five input neurons \cite{kriener2021yin, goeltz2021fast}.
As shown in \Cref{fig:yinyang}B the point's values, $x_i$ and $y_i$, in each direction, as well as their inverse, $1-x_i$ and $1-y_i$, are scaled linearly to spike times in the time window $\left[t_\text{early}, t_\text{late}\right]$.
In addition, a bias spike shortly after $t_\text{early}$ is inserted to increase the network activity which is observed for the duration $T$.
A hidden layer constitutes 120 \gls{lif} neurons in our \gls{snn} on \gls{bss2} integrates the input events and projects spikes itself onto a readout layer of 3 \gls{li} neurons, each corresponding to one class, as indicated by \Cref{fig:yinyang}A.
This allows it to infer a class decision $k_i$ for sample $\boldsymbol{x}_i$ by interpreting the maximum membrane value of the output neurons $v_k$ over time as a score for the corresponding class, i.e.\ $k = \argmax_k(\max_tv_k(t))$.
For each sample, the \gls{snn} is emulated for $T=\SI{60}{\micro\second}$.
We use the cross-entropy loss on the max-over-time values $s_k = \max_t v_k$ as the objective function, and SuperSpike \cite{neftci2019surrogate} surrogate gradients for training.
In \texttt{hxtorch.snn}, the PyTorch model is defined as outlined in \Cref{fig:swflow}B.
The upper plot in \Cref{fig:yinyang}D exemplifies the output traces while inferring a single sample.
The membrane voltage of the output neuron corresponding to the sample's class has the maximum value over time.
The lower plot depicts the achieved accuracy of the \gls{snn} on \gls{bss2} over epochs;
it reaches about $94.63 \pm 0.7$\% (standard deviation over 15 seeds).

\begin{table}
\begin{tabular}{lrr}
\toprule
section & duration [s] & rel.\ duration [\%]\\
\midrule
network emulation duration & 0.3 & 1.0\\
additional hardware runtime & 7.5 & 26.3\\
additional back-end overhead & 4.7 & 16.5\\
data transform to PyTorch & 13.1 & 45.7\\
additional front-end overhead & 2.5 & 8.6\\
gradient calculation & 0.8 & 2.9\\
\midrule
total duration & 28.7 & 100.0\\
\bottomrule
\end{tabular}
\caption{\label{tab:yinyang_performance}%
Runtime performance of a training epoch of the Yin-Yang experiment consisting of 64 batches to 75 samples using a host computer with an AMD Ryzen 7 3800X \gls{cpu}.
While the emulated network duration falls two orders of magnitude behind the total training runtime, the majority of the additional hardware and back-end time consists of the \gls{adc} neuron membrane potential readout.
The data transformation to PyTorch tensors is dominated by interpolation of the sparse-in-time \gls{adc} data onto a dense time grid.
The additional front-end overhead and gradient calculation don't contribute significantly to the total training duration.
}
\end{table}

\section{Discussion}\label{sec:discussion}

We presented the library \texttt{hxtorch.snn} to describe \gls{snn} models and their emulation on \gls{bss2} in the PyTorch ecosystem.
It integrates the auto-differentiation capabilities of PyTorch, providing access to the same training methodology as for \glspl{ann}.
Separation of network construction and emulation allows taking full advantage of the emulation speed-up of \gls{bss2} and the description of feed-forward and recurrent network topologies.
The \gls{api} is designed for user-defined extension of neuron and synapse types, including customization of the backward pass while removing the need for expert knowledge of the hardware.
Moreover, we showcased the library by using it to classify the YinYang dataset~\cite{kriener2021yin}.

The Yin-Yang samples are encoded in spike events as proposed in \cite{kriener2021yin} and fed into an \gls{snn} with one hidden \gls{lif} layer followed by a \gls{li} readout layer.
When emulated on \gls{bss2}, the \gls{snn} achieves a classification accuracy of $94.63 \pm 0.7$\% when using a max-over-time loss.
Compared to an accuracy of $95.0 \pm 0.9$\% achieved on \gls{bss2} in~\cite{goeltz2021fast}, where the authors use spiking output neurons and an analytical solution for the gradient to optimize the model, our accuracy is only marginally lower.
This verifies the design choices of our machine learning layer \texttt{hxtorch.snn} and demonstrates the implementation successfully.

The source code for the demonstration experiment, \texttt{hxtorch.snn}, and the underlying open-source software stack is available online~\cite{brainscales2022yinyang,githubelectronicvisions2022}.
As part of the EBRAINS research infrastructure \cite{ebrains2022ri}, we operate \gls{bss2} as a service allowing researchers to interactively conduct experiments and research on the neuromorphic platform.
In \cite{brainscales2022documentation} we provide and maintain a collection of experiments as interactive learning material for the platform.

\Cref{tab:yinyang_performance} evaluates the experiment runtime overhead of the software.
The surrogate-gradient-based training is dominated by data transformations, mostly from membrane measurement data.
However, more data-sparse training algorithms, such as EventProp~\cite{wunderlich2021event}, can avoid much of this conversion overhead.
To further optimize performance, adding support for event-based numerical codes would also be beneficial, as it would allow us to avoid converting to a time grid.
Here, for example, a port of \texttt{hxtorch.snn} to jax~\cite{jax2022github} could offer advantages based on increased flexibility yielding opportunities for more efficient data structures and codes.

The backward functions currently assigned to each module represent the entire backward dynamics of the module over time.
Since this is a constraint for recurrent topologies, we aim to incorporate algorithms and interfaces that allow to either assign a single backward function to a defined sub-network (and forward in mock mode) or, for a defined time grid, assign each module a function representing one integration step and then build the backward graph by implicit time-unrolling in an outer loop (and forward in mock mode).
Additionally, we plan to provide support for online learning rules like the e-prop learning method~\cite{bellec2020recurrently} involving code generation for the embedded \gls{simd} microprocessors on \gls{bss2}.

While currently one \texttt{Instance} instance corresponds to a single hardware experiment, we want to support mixed networks that are only partially executed on hardware by implicitly identifying the sub-networks to run on \gls{bss2} and execute the remaining parts in software.
In addition, we will implement the ability to execute layers partially in order to allow the use of layer sizes that cannot be mapped to \gls{bss2} due to limited neuron resources.
This will enable the pipelined and parallel execution of large layers on multiple chips.
Partial functionality of our \texttt{hxtorch.snn} will be consolidated in Norse~\cite{pehle2021norse}, and will serve as the mock backend, provide spike encoding and decoding schemes, and helper functions.

Overall, \texttt{hxtorch.snn} is a convenient and flexible tool for implementing and training \gls{snn} models on \gls{bss2}, utilizing the auto-differentiation capabilities of PyTorch to enable machine-learning-inspired training methods.
With further optimization and the support for the upcoming multi-chip systems, it has the potential to greatly facilitate and accelerate machine-learning-inspired \gls{snn} research and experimentation.

\begin{acks}

The authors would like to thank the BrainScaleS software team for fruitful discussions on software design,
the participants of the BrainScaleS-2 user meeting for their continuous support,
especially Sebastian Billaudelle and Johannes Weis for sharing their hardware expertise,
and all other current and former members of the Electronic Vision(s) research group who contributed to the \acrlong{bss2} neuromorphic platform.

This work has received funding from
the EC Horizon 2020 Framework Programme
under grant agreements
785907 (HBP SGA2) %
and
945539 (HBP SGA3), %
the \foreignlanguage{ngerman}{Deutsche Forschungsgemeinschaft} (DFG, German Research Foundation) under Germany’s Excellence Strategy EXC 2181/1-390900948 (the Heidelberg STRUCTURES Excellence Cluster),
the German Federal Ministry of Education and Research under grant number 16ES1127 as part of the \foreignlanguage{ngerman}{\emph{Pilotinnovationswettbewerb `Energieeffizientes KI-System'}},
from the Manfred Stärk Foundation,
and from the \foreignlanguage{ngerman}{Lautenschläger-Forschungspreis} 2018 for Karlheinz Meier.
\end{acks}

\section*{Author Contributions}\label{sec:author_contributions}

We give contributions in the \textit{CRediT} (Contributor Roles Taxonomy) format:
\textbf{PS \& EA}: Conceptualization, visualization, methodology, software, resources, writing — original draft, writing — reviewing \& editing;
\textbf{LB}: Investigation, validation, visualization, writing — original draft, writing — reviewing \& editing;
\textbf{CM}: Methodology, software, resources, writing — original draft, writing — reviewing \& editing;
\textbf{CP}: Conceptualization, methodology, writing — original draft, writing — reviewing \& editing;
\textbf{EM}: Conceptualization, methodology, software, resources, writing — original draft, writing — reviewing \& editing, supervision;
\textbf{JS}: Supervision, funding acquisition, writing — reviewing \& editing.

\bibliographystyle{ACM-Reference-Format}
\bibliography{vision,local}

\end{document}